\documentclass[10pt,letterpaper]{article}
\usepackage{envrethink-arxiv}
\usepackage[T1]{fontenc}
\usepackage{lmodern}
\usepackage[numbers,square,comma,sort&compress]{natbib}
\usepackage{amsmath,amssymb}
\usepackage{url}
\usepackage{hyphenat}
\usepackage{xspace}
\usepackage{enumitem}
\usepackage{tabularx}
\usepackage{pdflscape}
\usepackage{longtable}
\usepackage{flafter}
\usepackage{tikz}
\usepackage{graphicx}
\usepackage{booktabs}
\usepackage{array}
\usepackage{colortbl}
\definecolor{tblgroup}{RGB}{230,240,252}
\definecolor{tblgain}{RGB}{0,150,70}
\definecolor{tblloss}{RGB}{215,40,40}
\newcommand{\tblgrouprow}[2]{\rowcolor{tblgroup}\multicolumn{#1}{c}{\textbf{#2}}\\}
\newcommand{\gain}[1]{\textcolor{tblgain}{\textbf{#1}}}

\usepackage{placeins}
\usepackage{hyperref}

\usetikzlibrary{arrows.meta,positioning,fit,shapes.geometric,calc}
\newcommand{\bfit}[1]{\textbf{\textit{#1}}}
\newcommand{\oursys}{\textnormal{\textsc{Env-Rethink}}\xspace}
\newcommand{\environmenthard}{\mbox{Environment-Hard}\xspace}

\title{Breaking the Environment Wall: A Unified Framework for Preparing and Evolving Agent-Native Environments}
\newcommand{\shortpapertitle}{Breaking the Environment Wall}
\newcommand{\preprintlogos}{%
  \raisebox{-0.5\height}{\includegraphics[width=0.32\linewidth]{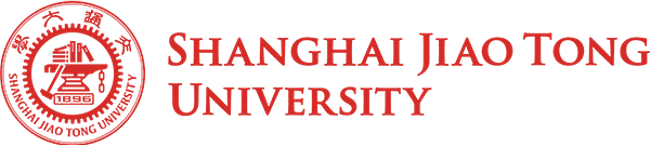}}\hfill
  \raisebox{-0.5\height}{\includegraphics[width=0.32\linewidth]{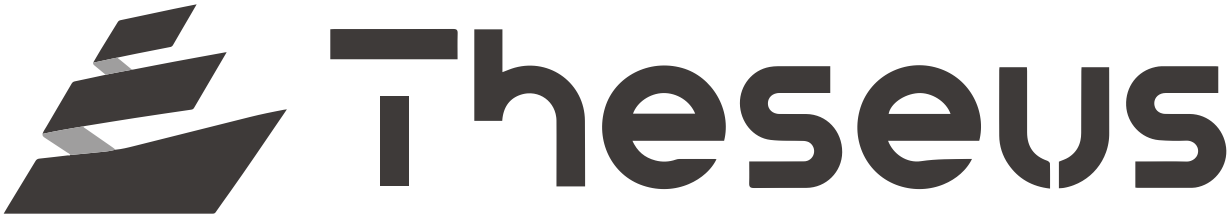}}%
}
\newcommand{\paperauthors}{%
  \mbox{Yukai Wu\textsuperscript{1,2,3}\thanks{Equal contribution.}}\quad
  \mbox{Yuanjing Yang\textsuperscript{1,2}\footnotemark[1]}\quad
  \mbox{Le Zhou\textsuperscript{1,2}}\quad
  \mbox{Shaokun Han\textsuperscript{1,2}}\quad
  \mbox{Haoyu Wang\textsuperscript{1,2}}\\[0.3em]
  \mbox{Zirui Tang\textsuperscript{1,2}}\quad
  \mbox{Xuzhou Zhu\textsuperscript{1,2}}\quad
  \mbox{Weihuang Zheng\textsuperscript{3}}\quad
  \mbox{Maxm Pan\textsuperscript{3}}\quad
  \mbox{Xuanhe Zhou\textsuperscript{1,2}\thanks{Corresponding author.}}\quad
  \mbox{Fan Wu\textsuperscript{1,2}}
}
\author{%
  \paperauthors\\[0.5em]
  {\small\textsuperscript{1}Shanghai Jiao Tong University\quad
  \textsuperscript{2}Theseus Labs\quad
  \textsuperscript{3}Tencent Hunyuan}%
}
\date{}
\hypersetup{
  hidelinks,
  pdftitle={Breaking the Environment Wall: A Unified Framework for Preparing and Evolving Agent-Native Environments},
  pdfauthor={Yukai Wu, Yuanjing Yang, Le Zhou, Shaokun Han, Haoyu Wang, Zirui Tang, Xuzhou Zhu, Weihuang Zheng, Maxm Pan, Xuanhe Zhou, Fan Wu}
}

\begin{document}
\maketitle
\begin{abstract}

Many real-world tasks (e.g., office workflows, scientific experimentation) require LLM agents to interact repeatedly with their environments for context-dependent operations. However, \textit{such environments are often not agent-ready}. First, information is often scattered and fragmented across the environment. 
Second, relevant evidence in the environment is often mixed with misleading information and conflicting versions. 
Third, environments evolve over time, introducing new noise and more challenging tasks. These challenges can substantially degrade performance for state-of-the-art AI agents (e.g., from {83.9\% to 57.6\%}).
To address these challenges, we propose \oursys (a system with 27B post-trained model) that supports three main capabilities: (1) It adaptively builds Collection Maps (for organizing related files) and Event Logs (for contextualizing cross-data relationships) to supplement necessary context; (2) It further leverages the post-trained model (through offline trajectory learning) to identify underlying noise issues in the environment; (3) It ultimately evolves environments through virtual event histories that alter environmental states and evidence relationships, producing more tricky ones for further agent improvement.
Experiments show that \oursys can effectively improve downstream task performance (with a {15.1 percentage-point increase in mean rubric pass rate} across nine models on 30 tasks). %

\end{abstract}

\begin{figure}[htbp]
\centering
\includegraphics[width=\linewidth]{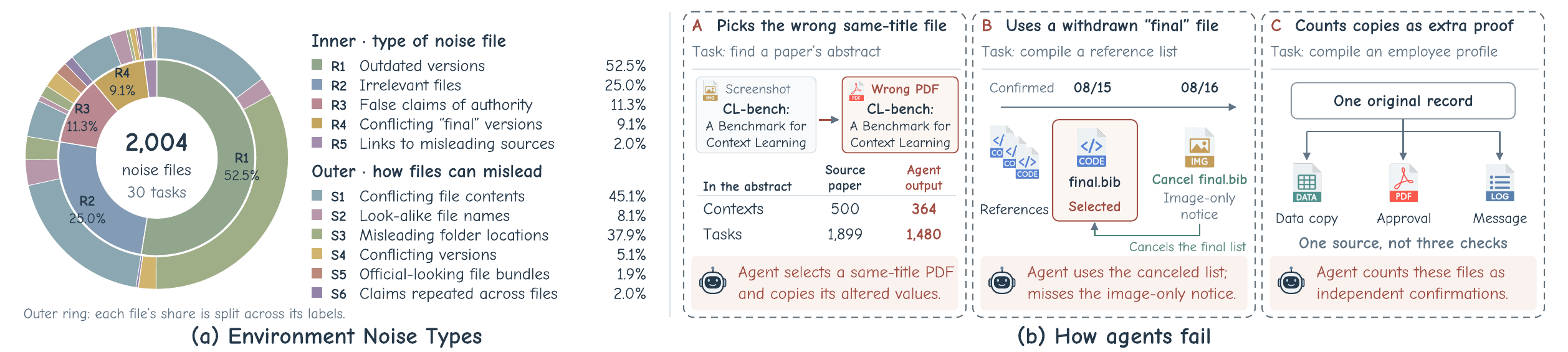}
\caption{{\textbf{Environmental Bottlenecks for AI Agents.} (a) Characteristics of 2,004 noise files across tasks from 30 real-world noisy environments. (b) Three typical AI agent failures upon environment-coupling tasks.}}
\label{fig:introduction-environment-wall}
\end{figure}

\clearpage
\FloatBarrier
\section{Introduction}
\label{stage:1-introduction}
\label{sec:introduction}
\label{stage:1-1-the-environment-wall}

{Real-world agents operate in persistent environments containing various artifacts, including code, data, configurations, documents, and records of prior actions. Frontier models such as Kimi K3 and increasingly capable harnesses support extended coding and multi-step workflows \citep{moonshot2026kimik3,anthropic2026harnessdesign}. However, these environments accumulate conflicting versions, misleading copies, and corrections that change which evidence remains valid. Strong task-solving capabilities alone do not resolve these relationships.}

{Current tools improve file access, context management, and continuity across sessions \citep{openai2025agentsdk,anthropic2025contextengineering,anthropic2025longrunning}. However, retrieving relevant-looking content does not establish its validity. Figure~\ref{fig:introduction-environment-wall}(b) illustrates three failures. GPT-5.6-Sol in Codex matches a PDF's title to the requested paper and checks its extracted statistics, yet copies altered values from the wrong source. In a reference-list task, it cross-checks earlier corrections but misses an image-only withdrawal and adopts the canceled list. In an employee-profile task, an agent mistakes records repeating one source for independent confirmation. {These failures, together with the need to support continued agent improvement, motivate three challenges.}}

\noindent$\bullet$ {\textbf{Recovering cross-file context.} A document and the evidence that invalidates it can differ in location, format, and time. Summarizing each file separately leaves this relationship implicit. The challenge is to connect related records while retaining the source evidence needed to interpret and verify their relationship.}

\noindent$\bullet$ {\textbf{Learning reliable evidence judgments.} The same title, version label, or confirmation can identify valid evidence in one environment and mislead in another. Even faithful extraction can therefore reproduce an incorrect source. The challenge is to learn transferable reading and verification decisions, beyond surface matching or task-specific rules.}

\noindent$\bullet$ {\textbf{Sustaining useful challenges as agents improve.} As agents become more capable, a fixed set of environments may offer fewer opportunities to expose new failures. Yet modifying an environment does not necessarily produce a useful learning challenge: changes may remove essential evidence or invalidate the expected answer and evaluation checks. The challenge is to introduce new reasoning demands while keeping the task solvable and its outcome verifiable.}

{There are some existing works that improve agent learning with environment-relevant design. EnvHarness reshapes interactions while preserving underlying environment logic and verifiers \citep{huang2026envharness}; RLVE adapts generated problem difficulty, and EvoEnv synthesizes executable, validated environments \citep{zeng2026rlve,shi2026evoenv}. Our focus is persistent file evidence: recovering its relationships, learning how to judge it, and changing its state together with the corresponding answers and checks.}

\label{stage:1-4-contributions-and-organization}
\noindent{\textbf{Our Methodology.} We present \oursys, which prepares and evolves agent environments through three complementary capabilities. In environment preparation stage, \oursys constructs a \emph{Collection Map} through semantic regrouping and an \emph{Event Log} through evidence-reviewed synthetic workflows. Additionally, \oursys conducts post-training on qualified tool-using trajectories for verifying file authority and version relationships. In environment evolution stage, \oursys creates plausible incorrect choices together with resolving evidence and updated evaluation assets. These capabilities support both reliable task execution and verifiable challenge generation.}

{We evaluate \oursys on tasks adapted from Workspace-Bench and on Terminal-Bench 2.1~{\citep{merrill2026terminalbench}}. Across nine downstream models, our 27B preparation model raises mean rubric pass rate from 57.6\% without preparation to 72.7\%, exceeding unadapted-model preparation by 13.3 percentage points. Held-out file-verification accuracy rises from 56.2\% to 76.5\%. Environment evolution lowers success for at least three of four models on 32 of 55 retained tasks with paired results under unchanged requests.}

\FloatBarrier

\Needspace{10\baselineskip}
\section{Preliminaries and Limitation Analysis}
\label{sec:preliminary}

\subsection{Agent-Native Environment}
\label{subsec:env-construction}

\noindent\textbf{Component Distinctions.} An agent system comprises a task model, a harness, and an environment. The task model $M_\theta$ performs reasoning and generates responses or tool-call requests; the harness $H$ manages context, executes tool calls, and controls the execution budget. Together, they form the task agent $A$, whose output for a request $q$ is $\hat{y}=A(q,E)$. \textit{The environment comprises the persistent resources and state that the agent accesses and manipulates through the harness.} We represent it as $E=(F,S)$, where $F$ contains files and service implementations and $S$ denotes their current state.

\begin{figure}[htbp]
    \centering
    \includegraphics[width=\linewidth]{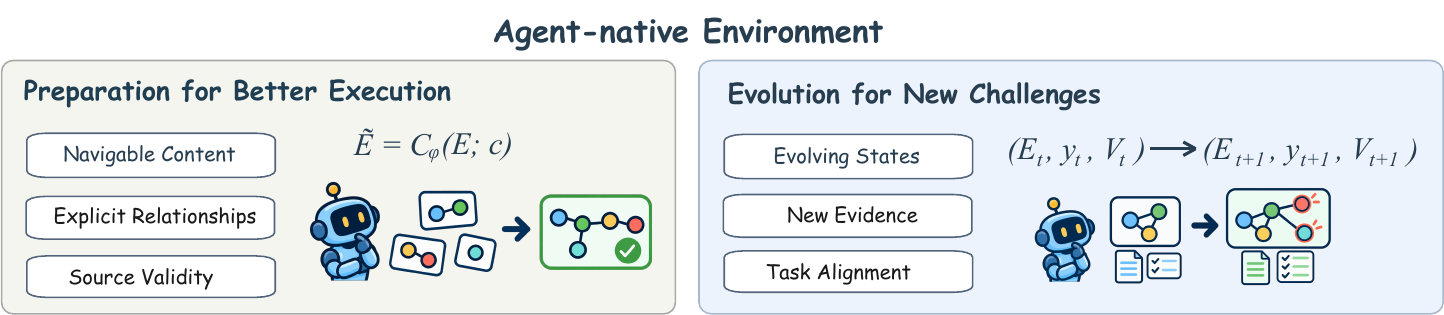}
    \caption{\textbf{Two complementary roles of an agent-native environment.} Preparation organizes existing evidence for task execution; evolution changes environmental states together with their reference outcomes and evaluators. Both preserve the task request.}
    \label{fig:agent-native-environment}
\end{figure}

Making an environment agent-native requires two complementary capabilities: \emph{environment preparation}, which makes existing evidence easier to locate and interpret for reliable execution, and \emph{environment evolution}, which creates new, verifiable challenges for continued learning. Both operate on the environment while preserving the task request, as illustrated in Figure~\ref{fig:agent-native-environment}.

\noindent$\bullet$ \bfit{Environment preparation} transforms an existing environment into $\widetilde{E}=C_\phi(E;c)$, where $C_\phi$ is a preparation operator and $c$ denotes optional auxiliary information, such as metadata or task-related supervision. It supports three properties: \emph{navigable content}, making relevant materials easy to locate; \emph{explicit relationships}, exposing how records support, revise, or depend on one another; and \emph{source validity}, providing inspectable evidence of authority and version applicability. Preparation supplies these properties through file selection, organization, and evidence-linked context, preserving the instance's correctness objective. The operator may use a trained preparation model without updating the downstream task model.

\noindent$\bullet$ \bfit{Environment evolution} changes environmental states to generate new, verifiable instances under the same request. It supports \emph{evolving states}, through changes to files, records, or applicable conditions; \emph{new evidence}, introducing further relationships to resolve; and \emph{task alignment}, keeping reference outcomes and evaluation checks consistent with the changed state. We express this process as
$(E_t,y_t,V_t)\rightarrow(E_{t+1},y_{t+1},V_{t+1})$,
where $y_t$ and $V_t$ denote the reference outcome and evaluator. The task agent receives only the resulting environment and unchanged request; reference outcomes and evaluation assets remain separate for validation. These instances provide new challenges and verification targets for subsequent model improvement.

\subsection{Limitations of Existing AI Agents}
\label{subsec:motivating-analysis}
\label{sec:motivating-pilot}\label{stage:1-2-motivating-experiment}

\begin{figure}[htbp]
    \centering
    \includegraphics[width=\linewidth]{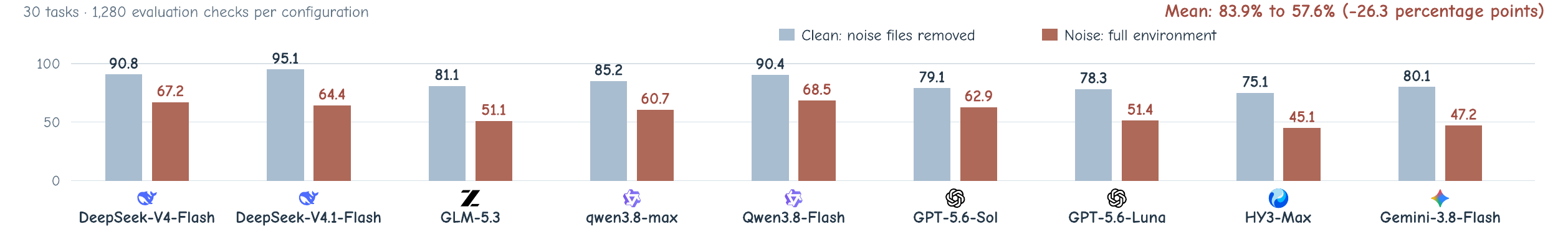}
    \caption{\textbf{Agent performance in clean vs. noisy environments.} Pass rates over 1,280 evaluation checks across 30 tasks for each of nine model-harness configurations. Clean uses ground-truth-selected files; Noise uses the full environment. Mean pass rate falls from 83.9\% to 57.6\%, a decrease of 26.3 percentage points.}
    \label{fig:clean-noise-performance}
\end{figure}

We evaluate nine model-harness configurations on 30 difficult tasks adapted from Workspace-Bench~\citep{tang2026workspacebench}, comparing ground-truth-selected files (\emph{Clean}) with all available files (\emph{Noise}) under fixed execution settings. Figure~\ref{fig:clean-noise-performance} shows that all nine configurations perform worse in noisy environments, with mean rubric pass rate falling from 83.9\% to 57.6\%.

\begin{figure}[!htbp]
    \centering
    \includegraphics[width=\textwidth]{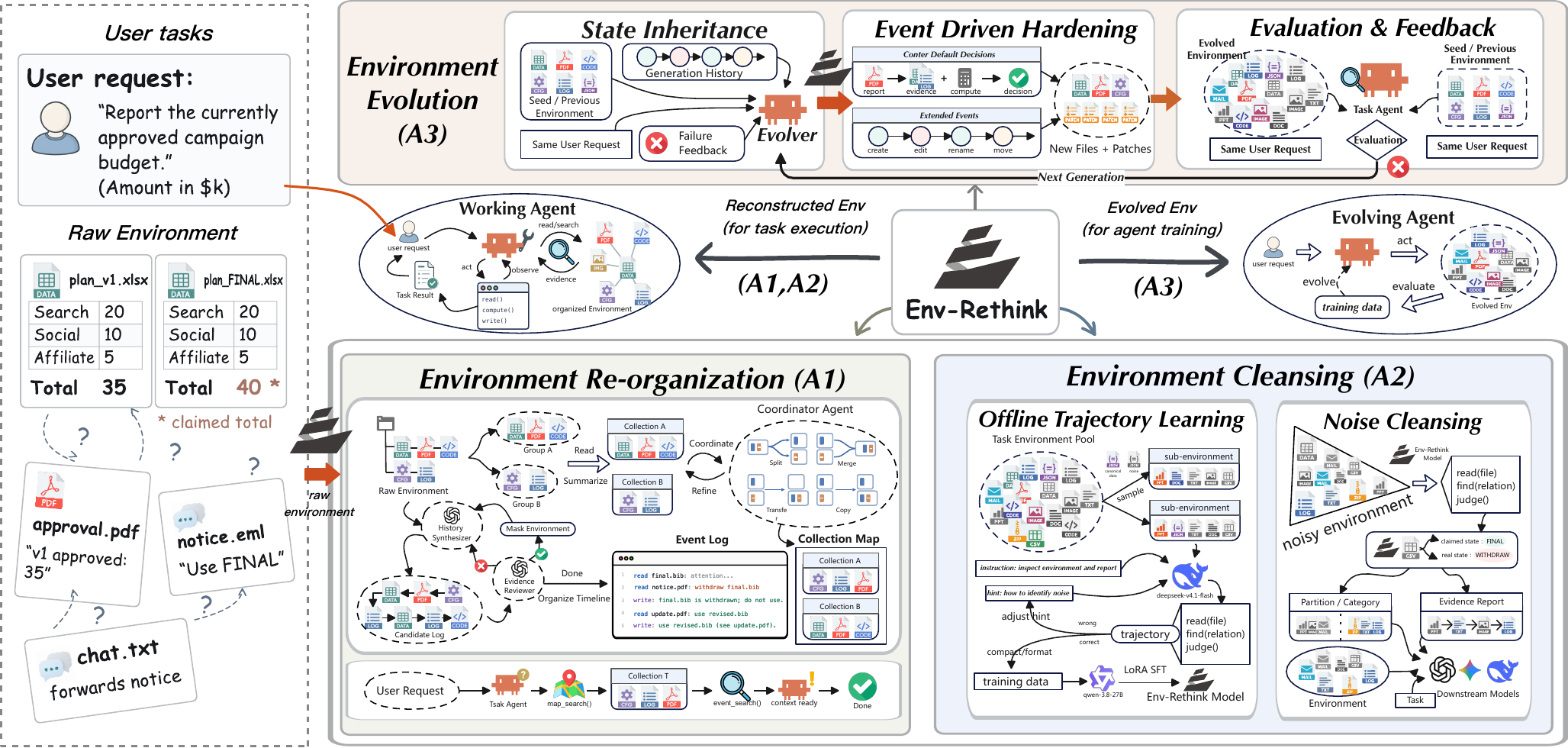}
    \caption{\textbf{Overview of \oursys.} Environment preparation combines reorganization (A1), which constructs Collection Maps and Event Logs, with cleansing (A2), which learns file verification from offline teacher trajectories and supplies selected evidence and a report to a downstream task agent. Environment evolution (A3) extends generation histories through event-driven changes and evaluates the resulting environments under the same user request. The budget example illustrates conflicting evidence in a raw environment.}
    \label{fig:overview}
    \label{fig:system-overview}
\end{figure}

Failure analysis reveals three recurring behaviors:
(1)~\emph{Incomplete search and reading}: agents overlook corrections, constraints, and supporting evidence.
(2)~\emph{Trusting filenames}: agents treat names such as \texttt{final.bib} as evidence of validity despite a later withdrawal notice. (3)~\emph{Unchecked reuse}: agents reuse similar files without verifying task requirements, carrying incompatible assumptions into their outputs.

\section{System Overview}
\label{sec:system-overview}
\label{sec:rsi-environment-overview}
\label{subsec:environment-stages}

{Figure~\ref{fig:system-overview} presents \oursys with three complementary components: \emph{environment reorganization} (A1) exposes cross-file context through Collection Maps and Event Logs; \emph{environment cleansing} (A2) learns file verification from qualified teacher trajectories and supplies selected evidence with a preprocessing report; and \emph{environment evolution} (A3) generates new states with updated reference outcomes and evaluation checks. These components support two agent workflows.}

\noindent{\underline{\textbf{$\blacktriangleright$ (1) Working Agent.}} The agent uses the environment prepared through reorganization and cleansing to locate related records, inspect evidence, and execute the user request. Its tool interactions produce the final task result.}

\noindent{\underline{\textbf{$\blacktriangleright$ (2) Evolving Agent.}} The agent executes the same request in evolved environments. Evaluation against updated reference outcomes and checks provides failure feedback for subsequent evolution, while validated environments supply new challenges for further agent training.}

\section{Methodology}
\label{sec:methodology}

\oursys addresses two complementary problems: preparing reliable evidence for task execution and generating harder, verifiable environments for continued learning. The preparation module combines \emph{reorganization}, which exposes cross-file context, with \emph{cleaning}, which learns to verify and select evidence. The evolution module changes environmental states together with their reference outcomes and evaluation checks while preserving the user request.

\subsection{Environment Preparation}
\label{stage:2-stage-1-task-ready-environment}

\noindent\textbf{Reorganization: exposing cross-file context.}
{The reference-list case in Figure~\ref{fig:introduction-environment-wall}(b) requires both locating the relevant records and establishing that the withdrawal invalidates the list. Directory proximity or separate summaries need not reveal this relationship. We combine coverage-preserving semantic regrouping (\emph{Collection Map}) with evidence-reviewed synthetic histories (\emph{Event Log}), without changing source files.}

{\emph{Semantic regrouping.} Deterministic grouping by directory and file count initializes collections. A \emph{Collection Summarizer} reads their files and writes cards with themes, contents, representative evidence, and source paths. The \emph{Semantic Coordinator} compares cards and member paths to propose splits, merges, transfers, or overlapping membership when supported by summaries or paths. Deterministic checks preserve every file in at least one collection. After each change, the \emph{Collection Refiner} rereads affected groups and updates the cards for the next round. This feedback keeps summaries consistent with membership as groups change beyond directory boundaries.}

{\emph{Evidence-reviewed histories.} The \emph{History Synthesizer} reads files and proposes reading, writing, copying, and exporting events linked to source evidence. Workflows and timestamps are marked synthetic; quotations remain distinct from inferred relationships. Deterministic checks validate event structure and file references, while an independent \emph{Evidence Reviewer} checks excerpts, actions, and relationships against the files. Feedback triggers repair; candidates exceeding the repair budget are discarded. The \emph{Timeline Integrator} merges accepted histories while preserving causal order. Thus, in the reference-list case, a later synthetic timestamp alone cannot establish withdrawal: the relationship requires supporting source evidence.}

{\emph{Inputs and execution.} Environment-Grounded Synthesis (EGS) uses only files and metadata. Task-Guided Synthesis (TGS) additionally uses clean/noise file labels for grouping and task requests and rubrics for history synthesis, simulating task-informed organization; instructions prohibit reproducing rubric entries in the public log. Both expose the same interface: the \emph{Task Executor} browses cards, retrieves sessions by file path, checks original evidence, and executes the request with native tools. Tool-use hooks also surface this context during execution.}

\noindent\textbf{Cleaning: learning evidence-based judgments.}
\label{stage:3-stage-2-agent-native-environment}
Organization alone cannot establish whether a purported final version is valid or an authority claim is supported. We therefore post-train a file verifier to inspect supporting and conflicting records. Given a collection, manifest, and reading tools, it predicts each file's standard/noise partition, category, observed stage, family, timeline, and supporting evidence. Separating the status claimed by a file from its status relative to other records discourages reliance on filenames or apparent authority. Verification excludes the downstream request; the task agent determines task relevance.

\noindent\textbf{Offline post-training.}
Independently sampled files may omit the evidence needed for verification. We instead construct sub-environments that retain authoritative records with their decoy chains, version families, and redirection evidence. Teachers inspect these collections through tools, producing trajectories of file access, observations, and judgments. Qualification checks verify reading fidelity, classification accuracy, retention of authoritative records, and version order. For failed trajectories, a separate agent uses annotations to revise the teacher's hint; unsuccessful retries are discarded. Student prompts omit these teacher hints.

We apply LoRA~{\citep{hu2022lora}} fine-tuning to Qwen3.8-27B using the qualified trajectories:
\begin{equation}
\mathcal{L}_{\mathrm{SFT}}(\theta)
=-\frac{\sum_{t=1}^{T}m_t\log p_\theta(x_t\mid x_{<t})}
{\sum_{t=1}^{T}m_t},
\label{eq:verification-sft}
\end{equation}
where $m_t$ selects assistant tokens, including tool calls and judgments; tool observations provide context without contributing to the loss. At deployment, verification produces selected files and a preprocessing report documenting the evidence and exclusions. A separate downstream agent executes the original request using this prepared environment.

\subsection{Environment Evolution}
\label{stage:4-stage-3-co-evolving-environment}
\label{sec:environment-evolving-agent}

\noindent\textbf{Creating answerable challenges.}
Changing files can introduce ambiguity or invalidate existing evaluation checks. We organize evolution around \emph{counter-default decision points}: each pairs a plausible incorrect choice with evidence that uniquely resolves it and a check that rejects it. These points concern effective dates, exclusions, exception rules, or numerical recalculation. Generation instructions require at least three points per variant and evidence chains no shorter than those in the seed.

\noindent\textbf{Joint state and reference updates.}
The evolver uses the current environment, generation history, and available failure feedback to produce file changes, updated reference outcomes, and revised checks under the unchanged request:
\[
(E_t,y_t,V_t)\rightarrow(E_{t+1},y_{t+1},V_{t+1}).
\]
File operations are recorded as events, allowing successive variants to build on earlier changes. A deterministic merger checks history continuity before assembling the new environment. This generation history, reference outcomes, and evaluation assets remain separate from the task agent's inputs.

\noindent\textbf{Validation and feedback.}
Structural checks verify assembly, references, evaluator syntax, and the unchanged request. Answer validation requires the updated reference solution to pass and the previous expected output to fail; shortcut audits check for leaked conclusions. Matched seed--variant runs then measure difficulty under identical agent configurations and budgets. Failed-check summaries guide subsequent evolution, while validated instances provide executable challenges and verification targets for further learning.

\section{Experiments}
\label{sec:experiments}

\subsection{Evaluation Overview}
\label{subsec:experimental-setup}

We evaluate environment organization, learned environment preparation, and environment
evolution in separate studies. The motivating Clean/Noise comparison appears in
Section~\ref{sec:motivating-pilot}. The following subsections retain each study's
inputs, evaluation conditions, and reported results; detailed training
and downstream evaluation settings appear in Appendix~\ref{sec:reproducibility},
and the evolution protocol appears in Appendix~\ref{stage:c-environment-protocol}.

We refer to the 30-task set in Section~\ref{sec:motivating-pilot} as \environmenthard.
The Clean and Noise settings are the GT and Full conditions in
Appendix~\ref{app:downstream-30tasks}. {Figure~\ref{fig:clean-noise-performance}}
shows that every configuration scores lower in Noise: rates range from 45.1\% to
68.5\%, compared with 75.1\% to 95.1\% in Clean. The mean difference is
26.3 percentage points.

\subsection{Environment Organization}
\label{stage:2-3-experimental-results}
\label{stage:2-3-1-office-workspace}
\label{sec:office-experiment}

\subsubsection{Comparing the Two Synthesis Modes}

Table~\ref{tab:evidence-summary} compares Task-Guided Synthesis (TGS),
{Environment-Grounded Synthesis (EGS)}, and the original noisy environment without the added
organization layer (Noise). Both synthesis modes provide Collection Map, Event Log,
and tool-use hooks that surface environment context during execution; TGS additionally
uses task-specific supervision during preparation.

\subsubsection{Synthesis Results}

The evaluation covers 30 tasks with 1,280 rubric items for every model. TGS outperforms
{EGS} for all five models. Both modes exceed the Noise baseline for all five models.
DeepSeek-V4.1-Flash improves from
64.4\% in Noise to 82.9\% with {EGS} and 87.3\% with TGS; GLM-5.3-Flash improves
from 58.3\% to 63.6\% and 81.6\%, respectively. For GPT-5.6-Sol, the complete
30-task comparison gives 78.2\% for TGS versus 69.6\% for {EGS}, a difference of
8.6 percentage points. Component ablations appear in Appendix~\ref{app:tgs-ablations}.

\begin{table}[htbp]
\centering
\caption{TGS, {EGS}, and Noise rubric pass rates (\%), with passed/total checks in
parentheses. TGS and {EGS} use the full environment layer (Collection Map, Event Log,
and tool-use hooks). All reported scores cover the same 30 tasks and 1,280 rubric items.
\textbf{Bold} denotes the best result in each column, and \underline{underlining} denotes the second best.
$\Delta$ reports the percentage-point difference between TGS and Noise.}
\label{tab:evidence-summary}
\label{tab:stage1-rerun}
\small
\setlength{\tabcolsep}{3pt}
\begin{tabular*}{\linewidth}{@{\extracolsep{\fill}}lrrrr@{}}
\toprule
\textbf{Model} & \textbf{TGS} & \textbf{{EGS}} & \textbf{Noise} & $\boldsymbol{\Delta}$ \\
\midrule
\tblgrouprow{5}{Harness: Claude Code}
DeepSeek-V4.1-Flash & \textbf{87.3} (1117/1280) & \textbf{82.9} (1061/1280) & \textbf{64.4} (824/1280) & \gain{+22.9} \\
GLM-5.3-Flash & \underline{81.6} (1044/1280) & 63.6 (814/1280) & 58.3 (746/1280) & \gain{+23.3} \\
Claude Opus 4.8 & 79.6 (1019/1280) & \underline{71.3} (913/1280) & 60.5 (774/1280) & \gain{+19.1} \\
HY3 & 62.0 (793/1280) & 46.3 (593/1280) & 45.1 (577/1280) & \gain{+16.9} \\
\midrule
\tblgrouprow{5}{Harness: Codex}
GPT-5.6-Sol & 78.2 (1001/1280) & 69.6 (891/1280) & \underline{62.9} (805/1280) & \gain{+15.3} \\
\bottomrule
\end{tabular*}
\end{table}

\subsubsection{From Context Synthesis to Environment-Understanding Training}

The two modes identify the next research question: how can a preparation model recognize valid
evidence from the environment's source files and reliably generate useful context? Task-guided
synthesis outperforms {EGS} in the matched comparisons, indicating that the
additional task-specific supervision improves the utility of the prepared context.
This motivates learning to identify relevant evidence and resolve file relationships
from materials in the environment.

Module 2 therefore turns these environmental judgments into training targets: acquiring
evidence, identifying source authority, resolving version relationships, and preparing
environments through file selection and structured reports, making environment
preparation learnable and reusable.

\subsection{Learned Environment Preparation}
\label{stage:3-2-training-evaluation}
\label{subsec:noise-evaluation}

We evaluate Qwen3.8-27B and the \oursys model on training-source and held-out task
environments, then measure downstream task performance using the prepared environments.

\subsubsection{Module Performance}
\label{stage:3-2-1-module-performance}

We evaluate Qwen3.8-27B and the \oursys model on the complete file collections of environments from 15
training-source tasks (1,579 files) and 15 held-out tasks (617 files). On training-source
environments, partition accuracy increases from 74.0\% to 90.0\%. On held-out
environments, it increases from 56.2\% to
76.5\%, and joint partition, category, and stage accuracy rises from 19.4\% to
32.4\%. Table~\ref{tab:module-evaluation} summarizes the comparison;
Appendix~\ref{app:module-evaluation-details} describes the task split and scoring protocol.

\begin{table}[!htb]
\centering
\small
\setlength{\tabcolsep}{4pt}
\caption{File-verification accuracy (\%) on training-source and held-out environments.
Each split contains 15 tasks, with 1,579 and 617 files, respectively. \textbf{Bold} denotes
\oursys. $\Delta$ reports the gain over Qwen3.8-27B in percentage points (pp).}
\label{tab:module-evaluation}
\label{fig:heldout-verification-results}
\label{fig:heldout-evaluation}
\begin{tabularx}{\linewidth}{@{}>{\raggedright\arraybackslash}Xrrrrrr@{}}
\toprule
& \multicolumn{3}{c}{\textbf{Training-source environments}}
& \multicolumn{3}{c}{\textbf{Held-out environments}} \\
\cmidrule(lr){2-4}\cmidrule(l){5-7}
\textbf{Metric} & Qwen3.8-27B & \oursys & $\Delta$
& Qwen3.8-27B & \oursys & $\Delta$ \\
\midrule
Partition accuracy & 74.0 & \textbf{90.0} & \gain{+16.0} & 56.2 & \textbf{76.5} & \gain{+20.3} \\
Noise-category accuracy & 39.8 & \textbf{56.8} & \gain{+17.0} & 33.4 & \textbf{38.1} & \gain{+4.7} \\
Observed-stage accuracy & 68.2 & \textbf{72.0} & \gain{+3.8} & 26.1 & \textbf{59.7} & \gain{+33.6} \\
Joint accuracy & 27.5 & \textbf{45.5} & \gain{+18.0} & 19.4 & \textbf{32.4} & \gain{+13.0} \\
\bottomrule
\end{tabularx}
\end{table}

\subsubsection{Downstream Utility}
\label{stage:3-2-2-downstream-utility}

We evaluate downstream utility on the 30-task \environmenthard set with nine downstream
models from the DeepSeek, GLM, Qwen, GPT, HY, and Gemini families. We compare three
environment conditions: \emph{Full environment} retains all noisy source files;
\emph{Qwen3.8-27B} uses the unadapted model to select files; and
\oursys uses the trained model for file selection and evidence reporting. Each model is evaluated
on the same 30 tasks under all three conditions.

Rubric pass rate pools passed checks over the 1,280 checks across these tasks.
Figure~\ref{fig:downstream-task-results} presents the comparison, with per-model
counts and aggregation details in Appendix~\ref{app:downstream-30tasks}.

\begin{figure}[!htb]
\centering
\includegraphics[width=\linewidth]{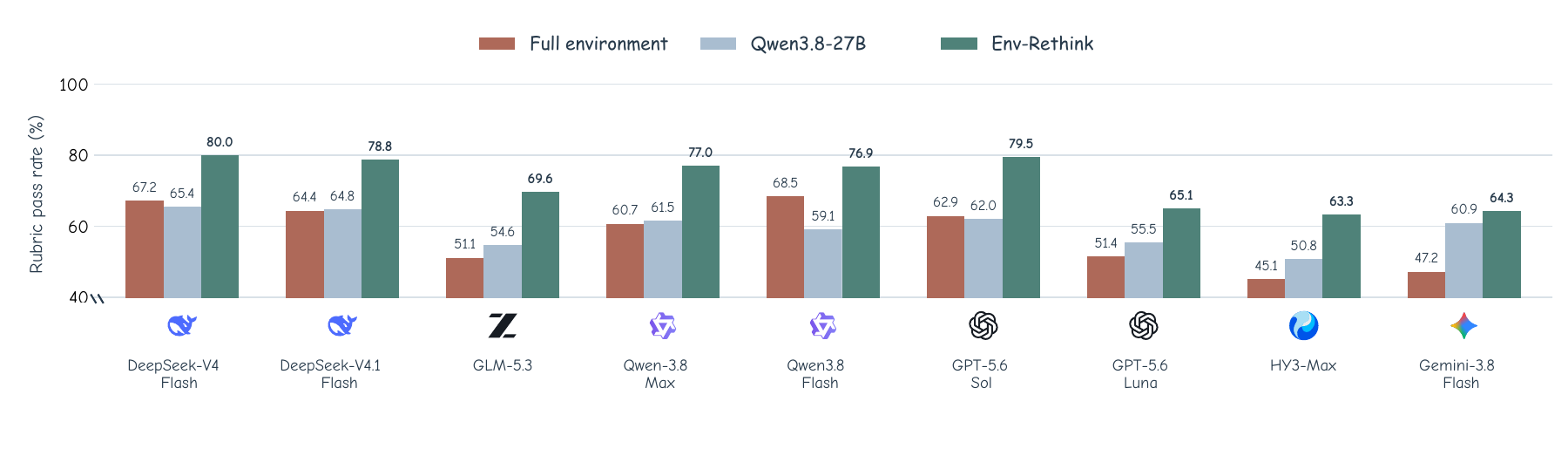}
\caption{Downstream rubric pass rates for nine models on the same 30 tasks.
The legend identifies the environment preparation condition; the horizontal axis
lists downstream task models. Each condition includes 1,280 checks per downstream
model. The vertical axis omits 0--40\%, as indicated by the axis break. %
}
\label{fig:downstream-task-results}
\end{figure}

Across the nine models, the mean pass rate rises from 57.6\% with full environments
and 59.4\% with Qwen3.8-27B to 72.7\% with \oursys.
\oursys improves over Qwen3.8-27B by 13.3 percentage points on average,
with gains across all nine models. Improvements over full environments range from
8.4 to 18.5 percentage points. These results show the downstream utility of the environments prepared by
\oursys through file selection and evidence reporting.

\subsubsection{Transfer and Reuse}
\label{stage:3-2-4-transfer-and-reuse}

The same \oursys model prepares environments for all nine downstream models across
the 30-task evaluation set. Its reuse occurs at the environment layer: file judgments
made during preparation determine the evidence available to each task agent. The
consistent improvements across downstream models show the utility of this shared
preparation step.

Module 2 produces a learned component that can be evaluated and versioned within an environment. Environment evolution uses virtual event histories to evolve seed states into harder instances under the
same task request, creating further examples for environment understanding and downstream work.

\subsection{Environment Evolution}
\label{stage:4-2-longitudinal-evaluation}
\label{subsec:generated-environment-evaluation}

We apply event-driven evolution to 59 Terminal-Bench 2.1~{\citep{merrill2026terminalbench}} seed tasks spanning
14 categories, including software engineering, debugging, model training, and data
processing. The evolved environments extend generation histories, materialize
changed states, and recompute reference outcomes while retaining the task requests.
We compare seed and evolved environments
using DeepSeek-V4.1-Flash, Gemini-3.8-Flash, Qwen3.8-Flash, and GLM-5.3-Flash,
with a nominal three trials per arm and native binary task-success scoring.
The analysis excludes a task with a failing variant reference run, leaving 55 tasks
with paired results.
Appendix~\ref{app:tb21-evaluation} gives the aggregation rules, task coverage,
and validation scope.

\begin{figure}[!htbp]
\centering
\includegraphics[width=\linewidth]{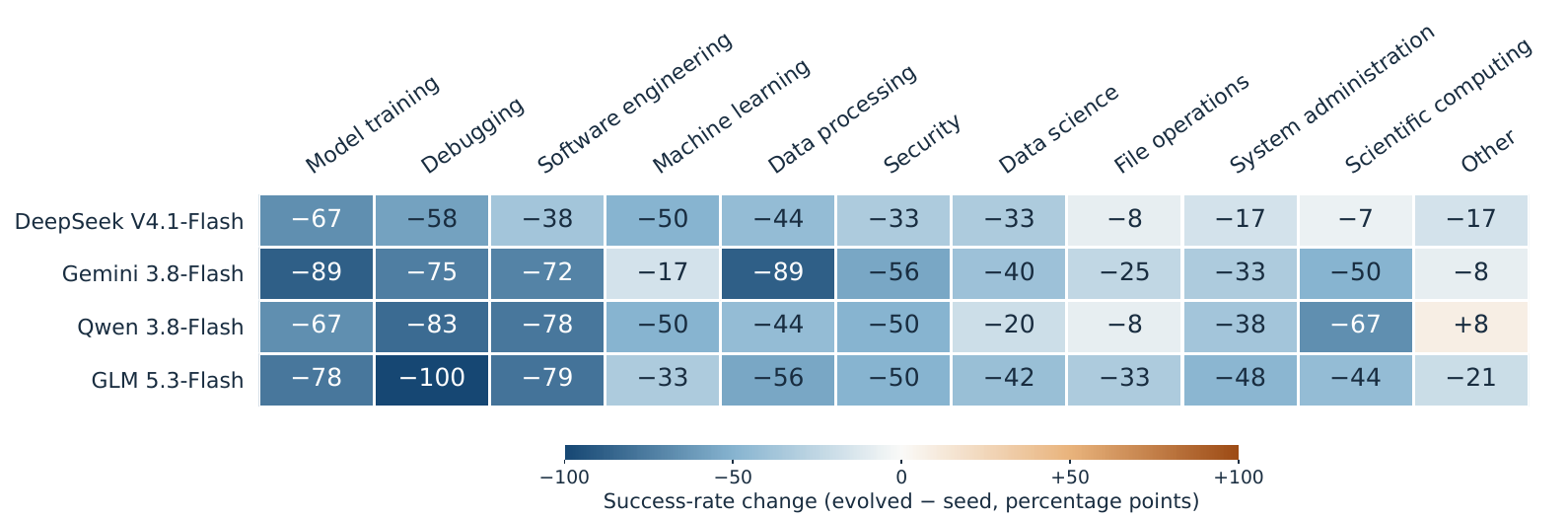}
\caption{\textbf{Environment evolution on Terminal-Bench 2.1.} Cell values give the reported
mean change in success rate (evolved minus seed, percentage points), with models
in rows and task categories in columns. Blue indicates lower success after evolution;
orange indicates higher success. Means use available pairs with nonzero seed success.
Other combines games, personal assistant, mathematics, and optimization. Evaluation
settings and validation details appear in Appendix~\ref{app:tb21-evaluation}.}
\label{fig:environment-evolution-heatmap}
\label{tab:stage3-evolution}
\end{figure}

\paragraph{Harder environments across tasks and models.}
Of the 55 retained tasks with paired results, {32 (58.2\%) show lower success in at
least three models}, including 21 with decreases in all four models. Four further
tasks show decreases in exactly two models. These effects span ten categories:
10 of 13 software-engineering tasks, four of five debugging tasks, and all three
model-training tasks meet the three-model criterion
(Figure~\ref{fig:environment-evolution-heatmap}). The category-level changes show that evolution
can expose new execution failures across distinct technical domains under the
same task requests.

\begin{samepage}
Environment organization and learned preparation help agents recover valid evidence;
environment evolution creates new situations in which that ability is tested.
The preparation results also reveal a useful balance between evidence retention and
noise filtering (Table~\ref{tab:constructor-quality}). Across the 30 environments,
\oursys retains 160 of 192 standard files, compared with 136 for Qwen3.8-27B,
while reducing admitted noise files from 606 to 271. The resulting workspaces
contain 431 files rather than 742, giving downstream agents a smaller collection
to inspect while preserving more standard evidence. Together,
these studies support environment preparation and challenge generation as resources
for continued improvement. Execution failures can inform subsequent generation,
and regenerated reference outcomes supply verification targets. This provides the challenge-generation component of
environment-driven recursive self-improvement.

\end{samepage}

\section{Related Work}
\label{stage:5-related-work}
\label{sec:related-work}

{\textbf{Agent tools and context management.} Agent interfaces determine what information can be found and carried across sessions. OpenAI provides file search, computer use, orchestration, and execution tracing \citep{openai2025agentsdk}. Anthropic uses selective retrieval, persistent notes, and structured handoffs \citep{anthropic2025contextengineering,anthropic2025longrunning,anthropic2026harnessdesign}, while SWE-agent designs interfaces for repository navigation and editing \citep{yang2024sweagent}. Together, these systems make information access and continuity part of agent execution.}

{\textbf{Environment adaptation and generation.} Constructed environments define the interactions available to agents and the outcomes that can be checked. EnvHarness adapts existing environments while preserving their logic and verifiers \citep{huang2026envharness}. RLVE and AgentScaler generate training tasks and tool-use simulations \citep{zeng2026rlve,fang2026agentscaler}, while AWM and EvoEnv produce stateful or executable environments \citep{wang2026awm,shi2026evoenv}. Across these approaches, environment structure and verifiability are explicit design choices.}

{\textbf{Learning from interaction.} Interaction records can turn observations and corrections into reusable guidance. ReAct grounds decisions in tool observations, and Reflexion retains textual feedback for later trials \citep{yao2023react,shinn2023reflexion}. OpenAI and Thrive Holdings' Tax AI turns practitioner corrections into evaluated fixes \citep{srinivasan2026taxai}; Factory's Signals identifies recurring failures across agent sessions and proposes code changes \citep{factoryresearch2026signals}. The value of a trace depends on how it informs later decisions.}

{\textbf{Feedback-guided environments.} Agent performance can guide which challenges an environment presents next. POET generates challenges alongside solution optimization and transfers solutions across environments \citep{wang2019poet}. PAIRED uses differences in agent returns to generate solvable challenges \citep{dennis2020paired}. Both link environment difficulty to observed learner behavior.}

\FloatBarrier

\section{Conclusion}
\label{stage:6-conclusion}
\label{sec:conclusion}

In this paper we studied how fragmented and unreliable evidence limits LLM agents in persistent file-based environments. We proposed Env-Rethink to prepare these environments for reliable execution and evolve them into harder, verifiable challenges. Env-Rethink supported (1) Collection Maps and evidence-reviewed Event Logs for organizing cross-file context, (2) a 27B model post-trained on tool-use trajectories for verifying and selecting evidence, and (3) event-driven evolution that updates environmental states, reference answers, and evaluation checks under unchanged requests. Empirically, learned preparation improved held-out file-partition accuracy from 56.2\% to 76.5\% and mean rubric pass rate from 57.6\% to 72.7\% across nine models on 30 tasks. Evolution reduced success for at least three of four models on 32 of 55 paired Terminal-Bench tasks. These results support environment preparation and evolution as complementary foundations for more reliable agents and future recursive self-improvement.

\FloatBarrier
\section*{AI Use Statement}
We used AI tools in three ways. First, language models are core components of
\oursys for environment organization, file verification, and event-driven evolution;
we fine-tune Qwen3.8-27B as the environment-preparation model. Second, models from
the DeepSeek, GLM, and Gemini families generate teacher trajectories, and AI tools
assist with file annotation and synthetic environment construction. These outputs
undergo the qualification, consistency checks, and human annotation review described
in the methodology and appendices. Third, AI tools assist with manuscript drafting
and editing, figure preparation, and literature and reference checks. The authors take responsibility
for the final content of this paper, including its text, claims, and artifacts.

\section*{Ethics Statement}
This work studies agent reliability in environments that simulate the complexity
of real-world workspaces, including fragmented information, conflicting versions,
and outdated records. Experiments use tasks adapted from Workspace-Bench and
Terminal-Bench. Constructed workflows and event histories are identified as
simulated and linked to source evidence. Data annotation includes consistency
checks and human review of challenging cases and disagreements. These procedures
support traceable environment construction and evaluation of agents' ability to
identify and use reliable evidence.

\section*{Reproducibility Statement}
Our anonymous repository is available at
\url{https://anonymous.4open.science/r/Env-Rethink-EDAD/}.
The methodology describes environment organization, trajectory qualification,
model adaptation, and event-driven evolution. Appendix~\ref{stage:a-data}
documents data selection, annotation definitions, and sub-environment construction.
Appendix~\ref{sec:reproducibility} provides training hyperparameters, hardware,
trajectory preparation, evaluation splits, scoring rules, and per-model downstream
results. Appendix~\ref{stage:c-environment-protocol} specifies the evolution
protocol, task coverage, and aggregation rules. Together, these details describe
how the reported environments, training corpus, and evaluation results are constructed.

\bibliographystyle{IEEEtran}
\bibliography{paper}

@inproceedings{shinn2023reflexion,
  author    = {Shinn, Noah and Cassano, Federico and Gopinath, Ashwin and Narasimhan, Karthik and Yao, Shunyu},
  title     = {Reflexion: Language Agents with Verbal Reinforcement Learning},
  booktitle = {Advances in Neural Information Processing Systems},
  volume    = {36},
  pages     = {8634--8652},
  publisher = {Curran Associates, Inc.},
  year      = {2023},
  url       = {https://proceedings.neurips.cc/paper_files/paper/2023/hash/1b44b878bb782e6954cd888628510e90-Abstract-Conference.html}
}

@inproceedings{wang2026awm,
  author    = {Wang, Zhaoyang and Xu, Canwen and Liu, Boyi and Wang, Yite and Han, Siwei and Yao, Zhewei and Yao, Huaxiu and He, Yuxiong},
  title     = {Agent World Model: Infinity Synthetic Environments for Agentic Reinforcement Learning},
  booktitle = {Proceedings of the 43rd International Conference on Machine Learning},
  year      = {2026},
  eprint    = {2602.10090},
  archivePrefix = {arXiv},
  primaryClass  = {cs.AI},
  url       = {https://arxiv.org/abs/2602.10090}
}

@inproceedings{wang2019poet,
  author    = {Wang, Rui and Lehman, Joel and Clune, Jeff and Stanley, Kenneth O.},
  title     = {{POET}: Open-Ended Coevolution of Environments and Their Optimized Solutions},
  booktitle = {Proceedings of the Genetic and Evolutionary Computation Conference},
  pages     = {142--151},
  publisher = {Association for Computing Machinery},
  address   = {New York, NY, USA},
  year      = {2019},
  doi       = {10.1145/3321707.3321799},
  url       = {https://doi.org/10.1145/3321707.3321799}
}

@inproceedings{dennis2020paired,
  author    = {Dennis, Michael and Jaques, Natasha and Vinitsky, Eugene and Bayen, Alexandre and Russell, Stuart J. and Critch, Andrew and Levine, Sergey},
  title     = {Emergent Complexity and Zero-Shot Transfer via Unsupervised Environment Design},
  booktitle = {Advances in Neural Information Processing Systems},
  volume    = {33},
  pages     = {13049--13061},
  publisher = {Curran Associates, Inc.},
  year      = {2020},
  url       = {https://proceedings.neurips.cc/paper/2020/hash/985e9a46e10005356bbaf194249f6856-Abstract.html}
}

@inproceedings{yang2024sweagent,
  author    = {Yang, John and Jimenez, Carlos E. and Wettig, Alexander and Lieret, Kilian and Yao, Shunyu and Narasimhan, Karthik and Press, Ofir},
  title     = {{SWE-agent}: Agent--Computer Interfaces Enable Automated Software Engineering},
  booktitle = {Advances in Neural Information Processing Systems},
  editor    = {Globerson, Amir and Mackey, Lester and Belgrave, Danielle and Fan, Angela and Paquet, Ulrich and Tomczak, Jakub and Zhang, Cheng},
  volume    = {37},
  pages     = {50528--50652},
  publisher = {Curran Associates, Inc.},
  year      = {2024},
  doi       = {10.52202/079017-1601},
  url       = {https://proceedings.neurips.cc/paper_files/paper/2024/hash/5a7c947568c1b1328ccc5230172e1e7c-Abstract-Conference.html}
}

@article{tang2026workspacebench,
  author        = {Tang, Zirui and Zhou, Xuanhe and Liu, Yumou and Li, Linchun and Wu, Yukai and Wang, Weizheng and Huang, Hongzhang and Zhou, Wei and Zhou, Jun and Song, Jiachen and Yu, Shaoli and Wang, Jinqi and Zhou, Zihang and Zhou, Hongyi and Lv, Yuting and Li, Jinyang and Liu, Jiashuo and Chen, Ruoyu and Liu, Chunwei and Li, GuoLiang and Kang, Jihua and Wu, Fan},
  title         = {{Workspace-Bench} 1.0: Benchmarking {AI} Agents on Workspace Tasks with Large-Scale File Dependencies},
  journal       = {arXiv preprint arXiv:2605.03596},
  year          = {2026},
  eprint        = {2605.03596},
  archiveprefix = {arXiv},
  primaryclass  = {cs.AI},
  doi           = {10.48550/arXiv.2605.03596},
  url           = {https://arxiv.org/abs/2605.03596},
  note          = {Preprint}
}

@inproceedings{yao2023react,
  author    = {Yao, Shunyu and Zhao, Jeffrey and Yu, Dian and Du, Nan and Shafran, Izhak and Narasimhan, Karthik and Cao, Yuan},
  title     = {{ReAct}: Synergizing Reasoning and Acting in Language Models},
  booktitle = {International Conference on Learning Representations},
  year      = {2023},
  url       = {https://openreview.net/forum?id=WE_vluYUL-X}
}

@misc{anthropic2025longrunning,
  author       = {Young, Justin},
  title        = {Effective Harnesses for Long-Running Agents},
  howpublished = {Engineering report},
  month        = nov,
  year         = {2025},
  url          = {https://www.anthropic.com/engineering/effective-harnesses-for-long-running-agents},
  urldate      = {2026-07-18}
}

@misc{openai2025agentsdk,
  author       = {{OpenAI}},
  title        = {New Tools for Building Agents},
  howpublished = {Product and engineering announcement},
  month        = mar,
  year         = {2025},
  url          = {https://openai.com/index/new-tools-for-building-agents/},
  urldate      = {2026-07-18}
}

@inproceedings{zeng2026rlve,
  author    = {Zeng, Zhiyuan and Ivison, Hamish and Wang, Yiping and Yuan, Lifan and Li, Shuyue Stella and Ye, Zhuorui and Li, Siting and He, Jacqueline and Zhou, Runlong and Chen, Tong and Zhao, Chenyang and Tsvetkov, Yulia and Du, Simon Shaolei and Jaques, Natasha and Peng, Hao and Koh, Pang Wei and Hajishirzi, Hannaneh},
  title     = {{RLVE}: Scaling Up Reinforcement Learning for Language Models with Adaptive Verifiable Environments},
  booktitle = {Proceedings of the 43rd International Conference on Machine Learning},
  year      = {2026},
  eprint    = {2511.07317},
  archivePrefix = {arXiv},
  primaryClass  = {cs.CL},
  url       = {https://arxiv.org/abs/2511.07317}
}

@inproceedings{fang2026agentscaler,
  author    = {Fang, Runnan and Cai, Shihao and Li, Baixuan and Wu, Jialong and Li, Guangyu and Yin, Wenbiao and Wang, Xinyu and Wang, Xiaobin and Su, Liangcai and Zhang, Zhen and Wu, Shibin and Tao, Zhengwei and Jiang, Yong and Xie, Pengjun and Zhang, Ningyu and Huang, Fei and Zhang, Wentao and Zhou, Jingren},
  title     = {Towards General Agentic Intelligence via Environment Scaling},
  booktitle = {Findings of the Association for Computational Linguistics: ACL 2026},
  publisher = {Association for Computational Linguistics},
  address   = {San Diego, California, United States},
  pages     = {17610--17621},
  month     = jul,
  year      = {2026},
  doi       = {10.18653/v1/2026.findings-acl.872},
  eprint    = {2509.13311},
  archivePrefix = {arXiv},
  url       = {https://aclanthology.org/2026.findings-acl.872/}
}

@article{shi2026evoenv,
  author  = {Shi, Yucheng and Liang, Zhenwen and Panaganti, Kishan and Yu, Dian and Yu, Wenhao and Mi, Haitao},
  title   = {Learning to Build the Environment: Self-Evolving Reasoning {RL} via Verifiable Environment Synthesis},
  journal = {arXiv preprint arXiv:2605.14392},
  year    = {2026},
  eprint  = {2605.14392},
  archivePrefix = {arXiv},
  primaryClass  = {cs.AI},
  url     = {https://arxiv.org/abs/2605.14392}
}

@article{huang2026envharness,
  author  = {Huang, Chengsong and Wang, Zifeng and Han, Rujun and Yan, Jun and Chen, Yanfei and CuiZhu, Zoey and Jiang, Ke and Xia, Peng and Yu, Han and Zhuang, Yufan and Ming, Yifei and Pan, Jiaqi and Mishra, Bhavana Dalvi and Huang, Jiaxin and Gokturk, Burak and Pfister, Tomas and Lee, Chen-Yu},
  title   = {{EnvHarness}: Awakening Static Worlds for Agent Learning},
  journal = {arXiv preprint arXiv:2608.19880},
  year    = {2026},
  eprint  = {2608.19880},
  archivePrefix = {arXiv},
  primaryClass  = {cs.AI},
  url     = {https://arxiv.org/abs/2608.19880}
}

@misc{moonshot2026kimik3,
 author = {{Moonshot AI}},
 title = {{Kimi K3}: Open Frontier Intelligence},
 year = {2026},
 howpublished = {Technical blog},
 url = {https://www.kimi.com/en/blog/kimi-k3},
 urldate = {2026-09-23}
}

@misc{anthropic2025contextengineering,
 author = {Rajasekaran, Prithvi and Dixon, Ethan and Ryan, Carly and Hadfield, Jeremy},
 title = {Effective Context Engineering for {AI} Agents},
 year = {2025},
 month = sep,
 howpublished = {Engineering report},
 url = {https://www.anthropic.com/engineering/effective-context-engineering-for-ai-agents},
 urldate = {2026-09-23}
}

@misc{anthropic2026harnessdesign,
 author = {Rajasekaran, Prithvi},
 title = {Harness Design for Long-Running Application Development},
 year = {2026},
 month = mar,
 howpublished = {Anthropic Engineering},
 url = {https://www.anthropic.com/engineering/harness-design-long-running-apps},
 urldate = {2026-09-23}
}

@misc{srinivasan2026taxai,
  author = {Srinivasan, Aravind and Shamdasani, Samay and Fernandes Araujo, Arthur and de Wasseige, John},
  title = {Building Self-Improving Tax Agents with Codex},
  year = {2026},
  month = may,
  howpublished = {OpenAI},
  url = {https://openai.com/index/building-self-improving-tax-agents-with-codex/}
}

@misc{factoryresearch2026signals,
  author = {{Factory Research}},
  title = {Signals: Toward a Self-Improving Agent},
  year = {2026},
  month = jan,
  howpublished = {Factory},
  url = {https://factory.com/news/factory-signals}
}

@misc{merrill2026terminalbench,
      title={{Terminal-Bench}: Benchmarking Agents on Hard, Realistic Tasks in Command Line Interfaces},
      author={Mike A. Merrill and Alexander G. Shaw and Nicholas Carlini and Boxuan Li and Harsh Raj and Ivan Bercovich and Lin Shi and Jeong Yeon Shin and Thomas Walshe and E. Kelly Buchanan and Junhong Shen and Guanghao Ye and Haowei Lin and Jason Poulos and Maoyu Wang and Marianna Nezhurina and Jenia Jitsev and Di Lu and Orfeas Menis Mastromichalakis and Zhiwei Xu and Zizhao Chen and Yue Liu and Robert Zhang and Leon Liangyu Chen and Anurag Kashyap and Jan-Lucas Uslu and Jeffrey Li and Jianbo Wu and Minghao Yan and Song Bian and Vedang Sharma and Ke Sun and Steven Dillmann and Akshay Anand and Andrew Lanpouthakoun and Bardia Koopah and Changran Hu and Etash Guha and Gabriel H. S. Dreiman and Jiacheng Zhu and Karl Krauth and Li Zhong and Niklas Muennighoff and Robert Amanfu and Shangyin Tan and Shreyas Pimpalgaonkar and Tushar Aggarwal and Xiangning Lin and Xin Lan and Xuandong Zhao and Yiqing Liang and Yuanli Wang and Zilong Wang and Changzhi Zhou and David Heineman and Hange Liu and Harsh Trivedi and John Yang and Junhong Lin and Manish Shetty and Michael Yang and Nabil Omi and Negin Raoof and Shanda Li and Terry Yue Zhuo and Wuwei Lin and Yiwei Dai and Yuxin Wang and Wenhao Chai and Shang Zhou and Dariush Wahdany and Ziyu She and Jiaming Hu and Zhikang Dong and Yuxuan Zhu and Sasha Cui and Ahson Saiyed and Arinbjörn Kolbeinsson and Jesse Hu and Christopher Michael Rytting and Ryan Marten and Yixin Wang and Alex Dimakis and Andy Konwinski and Ludwig Schmidt},
      year={2026},
      eprint={2601.11868},
      archivePrefix={arXiv},
      primaryClass={cs.SE},
      url={https://arxiv.org/abs/2601.11868},
}

@inproceedings{hu2022lora,
  title     = {{LoRA}: Low-Rank Adaptation of Large Language Models},
  author    = {Hu, Edward J. and Shen, Yelong and Wallis, Phillip and Allen-Zhu, Zeyuan and Li, Yuanzhi and Wang, Shean and Wang, Lu and Chen, Weizhu},
  booktitle = {International Conference on Learning Representations},
  year      = {2022},
  url       = {https://openreview.net/forum?id=nZeVKeeFYf9}
}

\clearpage
\appendix
\FloatBarrier

\section{Data}
\label{stage:a-data}
\label{app:noise-training}

{This appendix specifies the annotation and collection procedures supporting the
file-verification module.}

\subsection{Label Definitions}
\label{stage:a-1-label-definitions}

Each annotation records partition, category, observed stage, family, and evidence. The six
categories describe a file's authority and relationships within the supplied collection.

\begingroup
\small
\setlength{\tabcolsep}{4pt}
\renewcommand{\arraystretch}{1.2}
\begin{longtable}{@{}>{\raggedright\arraybackslash}p{\dimexpr 0.30000\linewidth-1.0000\tabcolsep\relax}>{\raggedright\arraybackslash}p{\dimexpr 0.70000\linewidth-1.0000\tabcolsep\relax}@{}}
\caption{File-verification category definitions.}\label{tab:file-categories}\\
\toprule
\textbf{Category} & \textbf{Meaning} \\
\midrule
\endfirsthead
\multicolumn{2}{l}{\small\itshape Table~\thetable{} (continued)}\\
\toprule
\textbf{Category} & \textbf{Meaning} \\
\midrule
\endhead
\midrule
\multicolumn{2}{r}{\small\itshape Continued on next page}\\
\endfoot
\bottomrule
\endlastfoot
\emph{authoritative record} & The designated authoritative record, supported by source records, content consistency, or archive lineage. \\
\emph{false final version} & A purported final or newest version contradicted by the genuine version chain or its own contents. \\
\emph{fabricated authority} & Fabricated approvals, messages, or other authority claims lacking supporting records. \\
\emph{misleading redirection} & A pointer or instruction conflicting with annotated authority relationships or the timeline. \\
\emph{superseded version} & A replaced member of a version family that may still supply historical evidence. \\
\emph{out-of-scope file} & A normal file outside the collection's target scope; task relevance is assessed for the specific downstream task. \\
\end{longtable}
\endgroup

\subsection{Data Manifest}
\label{stage:a-2-data-manifest}

{The construction rule allows identical file sets at most twice;} file, family, and semantic overlap are
checked across sub\hyp{}environments. Repeated teacher trajectories are handled separately through
the retention procedure in Appendix~\ref{stage:3-1-5-trajectory-and-model-selection}.

\subsection{Selection}
\label{stage:a-3-selection}

Training selection starts from the 30-task, 1,280-rubric \environmenthard pool and favors
strong clean performance with substantial noise sensitivity. The record contains 12 core tasks with
1,235 files and four reserve tasks, totaling 16 tasks and 1,597 files. Task 161 is excluded
from the 15-task construction pool, which contains 1,579 files. The original 209
sub\hyp{}environments and 30 subsequently generated held-out environments form a collection of
239. Qualified teacher trajectories cover 139 environments; the 30 additional environments and
the 70-environment teacher-failure pool are excluded from SFT training. The full-environment module evaluation uses the remaining 15 tasks, with 617 labeled files,
from the 30-task set. These parent tasks are separate from the 15-task training pool.

Selection uses Clean \ensuremath{\geq}90\%, a reported Noise\ensuremath{-}Clean contrast of
\ensuremath{-}38 to \ensuremath{-}75 percentage points, and file-level evidence of the failure
mechanism.

\subsection{Annotation}
\label{stage:a-4-annotation}

Filename and path rules initialize stage and category; task-analysis reports supply strong-decoy
evidence; teacher labeling covers remaining files. The process then checks manifests and family
constraints. Strong decoys and disagreements receive human review, while other files receive
category-stratified checks. The construction record reports that all 1,597 labeled files passed
consistency checks.

\subsection{Sub-Environments}
\label{stage:a-5-sub-environments}

Attack scenes retain authoritative targets with decoys and fabricated supporting chains. Version
families retain related authoritative and superseded records with their order. Redirect groups retain
pointers, targets, and supporting messages. Fillers add density around these relationships.
Composition varies the number of authoritative records and checks whether visible evidence supports each
label.

Small batches contain 4--12 files, medium batches 8--30, and large batches at most 80. The specified
mixture is 45\%/35\%/20\%. Size, semantic composition, and prompt condition accompany each
generation and acceptance record.

\subsection{Teacher Reading}
\label{stage:a-6-teacher-reading}

Current generation uses DeepSeek, GLM, and Gemini teacher families through Claude Code in isolated
sandboxes. Complete messages retain
calls, observations, and final judgments. Read-fidelity checks connect each file judgment to
content-access records and supporting observations.

The generation specification records three hint levels: L0 supplies the verification task, L1 adds
category features and distinguishes claimed stage from verified category, and L2 supplies a checking
sequence. The current student training prompt contains the verification instruction, schema, and
reading requirements without discriminative guidance. Prompt provenance records which generation
conditions produced each accepted trajectory and how it entered the final training corpus.

For retries, a separate agent examines the failed teacher trajectory and the ground-truth
annotations to revise the hint. The teacher receives the revised hint and repeats the
verification task.

\subsection{Qualification and Outputs}
\label{stage:a-7-qualification-and-outputs}

After read fidelity passes, deterministic qualification requires partition F1 \ensuremath{\geq}0.9,
zero false rejections of authoritative records, correct classification of all strong decoys, and correct order
where a version family occurs. The strong-decoy categories are \emph{false final version},
\emph{fabricated authority}, and \emph{misleading redirection}. The generation specification permits at most
two additional retries with hints revised by the separate agent described in
Appendix~\ref{stage:a-6-teacher-reading}; remaining failures are retained for analysis.

The current record reports 2,116 qualified trajectories. The dataset links their messages and
qualification records to file, family, scene, annotation, and sub-environment manifests.

\subsection{Unseen-Task Extension}
\label{app:unseen-task-data}

The extension contains tasks 72, 75, 78, 85, 100, 146, 159, 171, and 266, none of which entered
training, and records annotations for 304 files. Six existing-noise tasks (85, 100, 146, 159, 171,
and 266) contribute 224 files: 70 standard files and 154 noise files across 42 version families.
The other three tasks
(72, 75, and 78) receive 68 generated noise files across 18 version families. The construction
record includes solvability and task-navigation audits for these tasks.

The extension tasks are included in the unified 30-task downstream evaluation in
Appendix~\ref{app:downstream-30tasks}.

\FloatBarrier

\section{Reproducibility}
\label{stage:b-reproducibility}
\label{sec:reproducibility}

\subsection{Training Configuration}
\label{app:training-configuration}
\label{stage:3-1-4-training-configuration}

The training run uses model-native tool-call serialization: tool calls are
encoded as text in assistant content so that the rendered training messages
preserve them. The \oursys model is obtained by adapting Qwen3.8-27B with LoRA.

\begingroup
\small
\setlength{\tabcolsep}{4pt}
\renewcommand{\arraystretch}{1.2}
\begin{longtable}{
  @{}
  >{\raggedright\arraybackslash}
  p{\dimexpr 0.25000\linewidth-1.0000\tabcolsep\relax}
  >{\raggedright\arraybackslash}
  p{\dimexpr 0.75000\linewidth-1.0000\tabcolsep\relax}
  @{}
}
\caption{Configuration of the completed supervised fine-tuning run.}
\label{tab:training-configuration}\\
\toprule
\textbf{Item} & \textbf{Current configuration} \\
\midrule
\endfirsthead

\multicolumn{2}{l}{\small\itshape Table~\thetable{} (continued)}\\
\toprule
\textbf{Item} & \textbf{Current configuration} \\
\midrule
\endhead

\midrule
\multicolumn{2}{r}{\small\itshape Continued on next page}\\
\endfoot

\bottomrule
\endlastfoot

Base model
& Qwen3.8-27B; 27B dense model; recorded context capacity 256k \\

Adaptation
& LoRA supervised fine-tuning on multi-turn tool trajectories \\

Epochs
& 3 \\

LoRA rank / alpha
& 64 / 128 \\

Target modules
& All linear layers \\

Training examples / development examples
& 1,780 / 201 \\

Learning rate
& $1\times10^{-4}$, cosine schedule \\

Numerical precision
& bf16 \\

Training context
& 32k \\

Hardware
& One machine with 8 H20 GPUs; DeepSpeed ZeRO-3 \\

Optimization steps
& 168 \\

Training time
& 9 h 44 min 46 s \\

\end{longtable}
\endgroup

\subsection{Optimization Diagnostics}
\label{app:optimization-diagnostics}

The run completed three epochs in 168 optimization steps. The final training
loss was 0.199 and training token accuracy was 93.6\%; the final development
loss was 0.197 and development token accuracy was 94.1\%. Training curves cover 168 steps, with development evaluation at
optimization steps 30, 60, 90, 120, 150, and 168.

\begin{figure}[htbp]
\centering
\includegraphics[width=\linewidth]
  {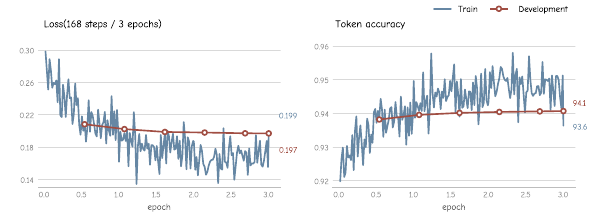}
\caption{Training and development loss and token accuracy over 168 steps
(3 epochs) of the fine-tuning run. Development evaluation occurs at
optimization steps 30, 60, 90, 120, 150, and 168. Final training loss/token
accuracy are 0.199/93.6\%, and final development values are 0.197/94.1\%.}
\label{fig:training-v2t-curves}
\label{fig:training-loss}
\label{fig:training-token-accuracy}
\end{figure}

The final evaluation is reported below.
Figure~\ref{fig:training-v2t-curves} shows the loss and token-accuracy curves.

\begingroup
\small
\setlength{\tabcolsep}{4pt}
\renewcommand{\arraystretch}{1.2}
\begin{table}[htbp]
\centering
\caption{Final optimization and development metrics.}
\label{tab:development-evaluations}
\begin{tabularx}{
  \linewidth
}{
  @{}
  >{\raggedright\arraybackslash}X
  >{\centering\arraybackslash}p{0.105\linewidth}
  >{\centering\arraybackslash}p{0.145\linewidth}
  >{\centering\arraybackslash}p{0.190\linewidth}
  >{\centering\arraybackslash}p{0.145\linewidth}
  >{\centering\arraybackslash}p{0.190\linewidth}
  @{}
}
\toprule
\textbf{Run}
& \textbf{Epochs}
& \textbf{Train loss}
& \textbf{Train token acc.}
& \textbf{Dev. loss}
& \textbf{Dev. token acc.} \\
\midrule
Final & 3 & 0.199 & 93.6\% & 0.197 & 94.1\% \\
\bottomrule
\end{tabularx}
\end{table}
\endgroup

\FloatBarrier

\subsection{Trajectory Preparation and Deployment}
\label{app:trajectory-preparation}
\label{stage:3-1-5-trajectory-and-model-selection}

Trajectory cleaning truncates only excessively long tool outputs while
preserving each invocation/response pair. Examples that remain overlength are
discarded as complete samples; training applies no further truncation. This
preparation retains complete conversational structure within the training
context. When a sub-environment has several teacher trajectories, a small
subset is retained by rotating among teachers, controlling repeated exposure
to the same environment while retaining variation in reading and reasoning.

The assembled corpus contains 1,780 training and 201 development examples. The
reported deployment merges the LoRA adapter into the base model after
optimization step 168. Module evaluation below reports results on full task environments.

\subsection{Module Evaluation Details}
\label{app:module-evaluation-details}

Table~\ref{tab:module-evaluation} evaluates the complete file collections of environments from both the 15 training-source
parent tasks and the 15 held-out parent tasks, using the updated file annotations. The 30-task
annotation set contains 2,196 files: 1,579 from the 15 training-pool tasks and 617
from the held-out tasks. We pool correct predictions over eligible files within each split.
For training-source environments, partition and joint accuracy use all 1,579 files,
noise-category accuracy uses 1,492 noise files, and observed-stage accuracy uses
1,328 files with a nonempty stage label. For held-out environments, partition
and joint accuracy use all 617 files, noise-category accuracy uses 512 noise files,
and observed-stage accuracy uses 372 files with a nonempty stage label. Missing
predictions count as incorrect. Joint accuracy requires the partition, category,
and stage judgments to agree with the reference labels.

On training-source environments, the Qwen3.8-27B and \oursys correct counts are respectively
1,169 and 1,421 for partition, 594 and 848 for noise category, 906 and 956 for stage,
and 434 and 719 for joint accuracy. On held-out environments, they are 347 and 472 for partition,
171 and 195 for noise category, 97 and 222 for stage, and 120 and 200 for joint
accuracy. All 15 held-out tasks contribute to these totals.

\FloatBarrier

\subsection{Environment Preparation Details}
\label{app:workspace-construction-details}

The preparation model selects files and supplies an evidence report for downstream use.
We evaluate selection quality on 30 task environments containing 2,196 files:
192 standard files and 2,004 noise files. Table~\ref{tab:constructor-quality}
reports retention and filtering quality. \oursys retains 160 of
192 standard files and admits 271 noise files, compared with 136 and 606,
respectively, for Qwen3.8-27B.

\begin{table}[htbp]
\centering
\small
\setlength{\tabcolsep}{4pt}
\caption{Environment-preparation quality over 30 tasks and 2,196 files.
Rejected standard files and admitted noise files are counts; recall is
measured over standard-labeled files.}
\label{tab:constructor-quality}
\begin{tabularx}{\linewidth}{@{}>{\raggedright\arraybackslash}Xrrrr@{}}
\toprule
\textbf{Preparation model}
& \textbf{Selected}
& \shortstack{\textbf{Standard}\\\textbf{rejected}}
& \shortstack{\textbf{Noise}\\\textbf{admitted}}
& \shortstack{\textbf{Standard}\\\textbf{recall}} \\
\midrule
\oursys & 431 & 32 & 271 & 83.3\% \\
Qwen3.8-27B      & 742 & 56 & 606 & 70.8\% \\
Ground truth     & 192 & 0  & 0   & 100.0\% \\
\bottomrule
\end{tabularx}
\end{table}

\FloatBarrier

\subsection{Thirty-Task Downstream Evaluation}
\label{app:downstream-30tasks}
\label{app:additional-reuse-results}
\label{app:unseen-task-evaluation}

The evaluation covers nine downstream models on the same 30 tasks. Full retains
all files in the original environment; Qwen3.8-27B uses files selected by the
unadapted model; \oursys uses files and an evidence report prepared by our trained
model. The model names identify environment preparation conditions. We also report a GT-curated environment reference, which
selects files using ground-truth labels.

\begin{table}[htbp]
\centering
\small
\setlength{\tabcolsep}{3pt}
\caption{Thirty-task downstream results. Full, Qwen3.8-27B, and \oursys report passed checks
out of 1,280 and the corresponding rate. GT gives the reference pass rate on
environments prepared using label-based file selection. \textbf{Bold} denotes \oursys, and
\underline{underlining} denotes the better of the two baselines (Full and Qwen3.8-27B).
$\Delta$ is the difference between \oursys and Qwen3.8-27B, in percentage points.}
\label{tab:downstream-conditions}
\label{tab:downstream-results}
\begin{tabularx}{\linewidth}{@{}>{\raggedright\arraybackslash}Xrrrrr@{}}
\toprule
& \multicolumn{2}{c}{\textbf{Baselines}} & & \textbf{Reference} & \\
\cmidrule(lr){2-3}\cmidrule(lr){5-5}
\textbf{Downstream model} & Full & Qwen3.8-27B & \textbf{\oursys} & GT & $\Delta$ \\
\midrule
DeepSeek-V4-Flash & \underline{860} (67.2\%) & 837 (65.4\%) & \textbf{1,024 (80.0\%)} & 90.8\% & \gain{+14.6} \\
DeepSeek-V4.1-Flash & 824 (64.4\%) & \underline{829} (64.8\%) & \textbf{1,008 (78.8\%)} & 95.1\% & \gain{+14.0} \\
GLM-5.3 & 654 (51.1\%) & \underline{699} (54.6\%) & \textbf{891 (69.6\%)} & 81.1\% & \gain{+15.0} \\
Qwen-3.8-Max & 777 (60.7\%) & \underline{787} (61.5\%) & \textbf{986 (77.0\%)} & 85.2\% & \gain{+15.5} \\
Qwen3.8-Flash & \underline{877} (68.5\%) & 756 (59.1\%) & \textbf{984 (76.9\%)} & 90.4\% & \gain{+17.8} \\
GPT-5.6-Sol & \underline{805} (62.9\%) & 793 (62.0\%) & \textbf{1,018 (79.5\%)} & 79.1\% & \gain{+17.6} \\
GPT-5.6-Luna & 658 (51.4\%) & \underline{710} (55.5\%) & \textbf{833 (65.1\%)} & 78.3\% & \gain{+9.6} \\
HY3-Max & 577 (45.1\%) & \underline{650} (50.8\%) & \textbf{810 (63.3\%)} & 75.1\% & \gain{+12.5} \\
Gemini-3.8-Flash & 604 (47.2\%) & \underline{780} (60.9\%) & \textbf{823 (64.3\%)} & 80.1\% & \gain{+3.4} \\
\midrule
Mean rate & 57.6\% & \underline{59.4\%} & \textbf{72.7\%} & -- & \gain{+13.3} \\
\bottomrule
\end{tabularx}
\end{table}

Per-model rates pool passed checks across tasks. Missing outputs count as zero
passed checks while retaining the task's rubric denominator. Each condition
therefore covers 11,520 checks across the nine models. Means are computed from
unrounded counts. The reported scores incorporate corrected evaluator judgments.
GT-curated scores serve as an empirical reference for each downstream model.

The task set is 72, 75, 78, 79, 85, 87, 94, 100, 108, 124, 129, 146, 154,
159, 160, 161, 171, 207, 258, 266, 267, 288, 291, 300, 314, 334, 357, 359,
372, and 374. The nine-task annotation extension in
Appendix~\ref{app:unseen-task-data} contributes to this evaluation set.

\FloatBarrier

\subsection{TGS Component Ablations}
\label{app:tgs-ablations}

On the same 30-task set, removing Collection Map reduces DeepSeek-V4.1-Flash's rubric pass
rate from 87.3\% to 75.1\% ($-12.2$ percentage points), while removing Event Log
reduces it to 84.8\% ($-2.4$ percentage points).
Both ablations retain the tool-use hooks and the remaining environment component.

\begin{table}[htbp]
\centering
\caption{DeepSeek-V4.1-Flash TGS component ablations on 30 tasks with 1,280 rubric
items. All conditions retain tool-use hooks. Changes relative to the full condition
are in percentage points, calculated from unrounded counts.}
\label{tab:tgs-ablations}
\begin{tabular}{lrrr}
\toprule
Condition & Passed/total & Rate (\%) & $\Delta$ vs. full \\
\midrule
Full (Collection Map + Event Log + hooks) & 1117/1280 & 87.3 & -- \\
w/o Collection Map & 961/1280 & 75.1 & $-12.2$ \\
w/o Event Log & 1086/1280 & 84.8 & $-2.4$ \\
\bottomrule
\end{tabular}
\end{table}
\FloatBarrier

\FloatBarrier

\section{Event-Driven Generation Protocol}
\label{stage:c-environment-protocol}

The materialized environment and its generation record are represented separately:
\begin{equation}
E_t=\langle F_t,S_t\rangle,\qquad
R_t=\langle L_t,y_t,V_t;\nu_t\rangle,
\label{eq:environment-state}
\end{equation}
where $F_t$ comprises files and service implementations, $S_t$ is their state,
$L_t$ is the generation history, $y_t$ is the reference outcome, $V_t$ is the
evaluator, and $\nu_t$ identifies the variant. The task agent receives $E_t$ and
the unchanged request $q$; generation and verification retain $R_t$.

\paragraph{Generation and validation.}
The Environment Evolver receives the seed or previous variant, its generation
history, and available execution feedback. It produces environmental changes,
causally linked events, and revised reference outcomes. Assembly applies the changes
to produce a runnable environment. Structural checks verify event links, file
references, evaluation assets, and byte-for-byte preservation of the task request.
Reference execution checks that the new answer passes and the previous answer fails.
Each variant specifies at least three counter-default decision points, linking a
plausible incorrect choice to resolving evidence and an executable check. Shortcut
audits inspect generated materials for conclusions that would bypass the seed's
reasoning requirements. Evaluation uses the assembled environment, preserving its
state when files are supplied to the agent.

\subsection{Evaluation Settings}
\label{app:tb21-evaluation}

\paragraph{Task coverage.}
The collection contains 59 Terminal-Bench 2.1 seeds: four easy and 55 medium tasks.
We exclude \texttt{sanitize-git-repo} because its variant reference run fails, leaving
58 seeds, of which 55 have paired results. Among these, 32 show a success-rate decrease
of at least 20 percentage points in three or more models, four in exactly two models,
and 19 in at most one model. The reported 58.2\% uses the 55 retained paired tasks as
its denominator. Category summaries apply the same exclusion.
{The aggregate reports observed success-rate changes.} A seed-environment fault is
recorded for \texttt{mteb-retrieve}.

\paragraph{Scoring and aggregation.}
We evaluate DeepSeek-V4.1-Flash, Gemini-3.8-Flash, Qwen3.8-Flash, and GLM-5.3-Flash
with matched configurations and execution budgets across seed and variant.
The target allocation is three independent trials per arm. Task success requires
passing all native checks; check-level pass rates are used only for diagnosis.
For task $i$ and model $m$,
\begin{equation}
\Delta_{i,m}=100\bigl(\hat p^{\mathrm{evolved}}_{i,m}
-\hat p^{\mathrm{seed}}_{i,m}\bigr),\qquad
h_i=\sum_m\mathbf{1}\{\Delta_{i,m}\leq -20\}.
\end{equation}
The main result counts tasks with $h_i\geq3$. This is a screening criterion:
one changed outcome in three trials corresponds to 33.3 percentage points.
Rates use completed-trial denominators; for example, GLM's seed score on
\texttt{largest-eigenval} is 1/2 and its variant score is 0/3.

{Category means in Figure~\ref{fig:environment-evolution-heatmap} average task-level
contrasts, excluding missing pairs and pairs with zero seed success.}
Other combines games, personal assistant, mathematics, and optimization.

\end{document}